\documentclass[letterpaper, 10pt, conference]{ieeeconf}

\IEEEoverridecommandlockouts %
\usepackage[space]{cite}
\usepackage{amsmath}
\usepackage{amssymb}
\usepackage{graphicx}
\usepackage{booktabs}
\usepackage{xcolor}
\usepackage{cleveref}
\let\labelindent\relax
\usepackage[inline]{enumitem}
\usepackage{stfloats}
\usepackage{tabularx}
\usepackage{array}
\usepackage{url}
\newcommand{\T}{^{\mathsf{T}}}
\graphicspath{{figs/}}

\title{\LARGE \bf
FORM: Robot Manipulation through Direct Material Law Identification
}

\author{Stepan Tretiakov$^{1,*,\dagger}$, Ruihan Zhao$^{2,*}$, Cheng-Hsi Hsiao$^{3}$, Xingjian Li$^{4}$, Adam Thorpe$^{4}$,\\
Hassan Iqbal$^{4}$, Sandeep Chinchali$^{2}$, Ufuk Topcu$^{5}$, Krishna Kumar$^{3,4}$%
\thanks{This work has been submitted to the IEEE for possible publication. Copyright may be transferred without notice, after which this version may no longer be accessible.}%
\thanks{$^{1}$Department of Mechanical Engineering,
University of California, Berkeley, CA 94720, USA.
$^{2}$Chandra Family Department of Electrical and Computer Engineering,
$^{3}$Maseeh Department of Civil, Architectural and Environmental Engineering,
$^{4}$Oden Institute for Computational Engineering and Sciences, and
$^{5}$Department of Aerospace Engineering and Engineering Mechanics,
The University of Texas at Austin, Austin, TX 78712, USA.}%
\thanks{\protect\raggedright Emails: stepa@berkeley.edu, ruihan.zhao@utexas.edu, chhsiao@utexas.edu,
xingjian.li@austin.utexas.edu, adam.thorpe@austin.utexas.edu, hassan.iqbal@utexas.edu,
sandeepc@utexas.edu, utopcu@utexas.edu, krishnak@utexas.edu}%
\thanks{$^{*}$Equal contribution.}%
\thanks{$^{\dagger}$Corresponding author.}%
}

\begin{document}

\maketitle
\thispagestyle{empty}
\pagestyle{empty}

\begin{abstract}
When interacting with an unfamiliar deformable material, a robot lacks prior knowledge of its physical properties and how it will respond to applied forces and motion. Rapid online identification is therefore essential for reliable manipulation. We present FORM (From Observed Response to Material laws), which identifies material properties from a single robot interaction and reuses the recovered model to plan manipulation under new actions and geometries. We use weak-form momentum balance to convert observed material motion and contact forces into linear equations in the unknown material parameters. These equations are assembled using the same material point method discretization as the forward simulator, so identification reduces to linear least-squares solves whose solutions can be used directly for prediction without refitting or conversion. Across four material classes, FORM reduces identification time from roughly 10–25 minutes for iterative baselines to 2–5 seconds, while maintaining competitive accuracy on new motions, initial conditions, and geometries. We demonstrate our approach in simulation and on hardware across four manipulation tasks: elastic rod insertion, golf putting with an elastic club, elastoplastic shaping, and target-volume pouring. In each task, the model identified from a single interaction is reused to plan new motions or manipulate a different geometry. FORM estimates elastic properties within 3.4\% and elastoplastic properties within 2\%, achieves 72.4–77.8\% IoU in dough shaping, and keeps mean pouring error at 3.8 mL across target volumes of 60–160 mL.
\end{abstract}

\section{Introduction}
\label{sec:intro}

\begin{figure}[t]
    \centering
    \includegraphics[width=\linewidth]{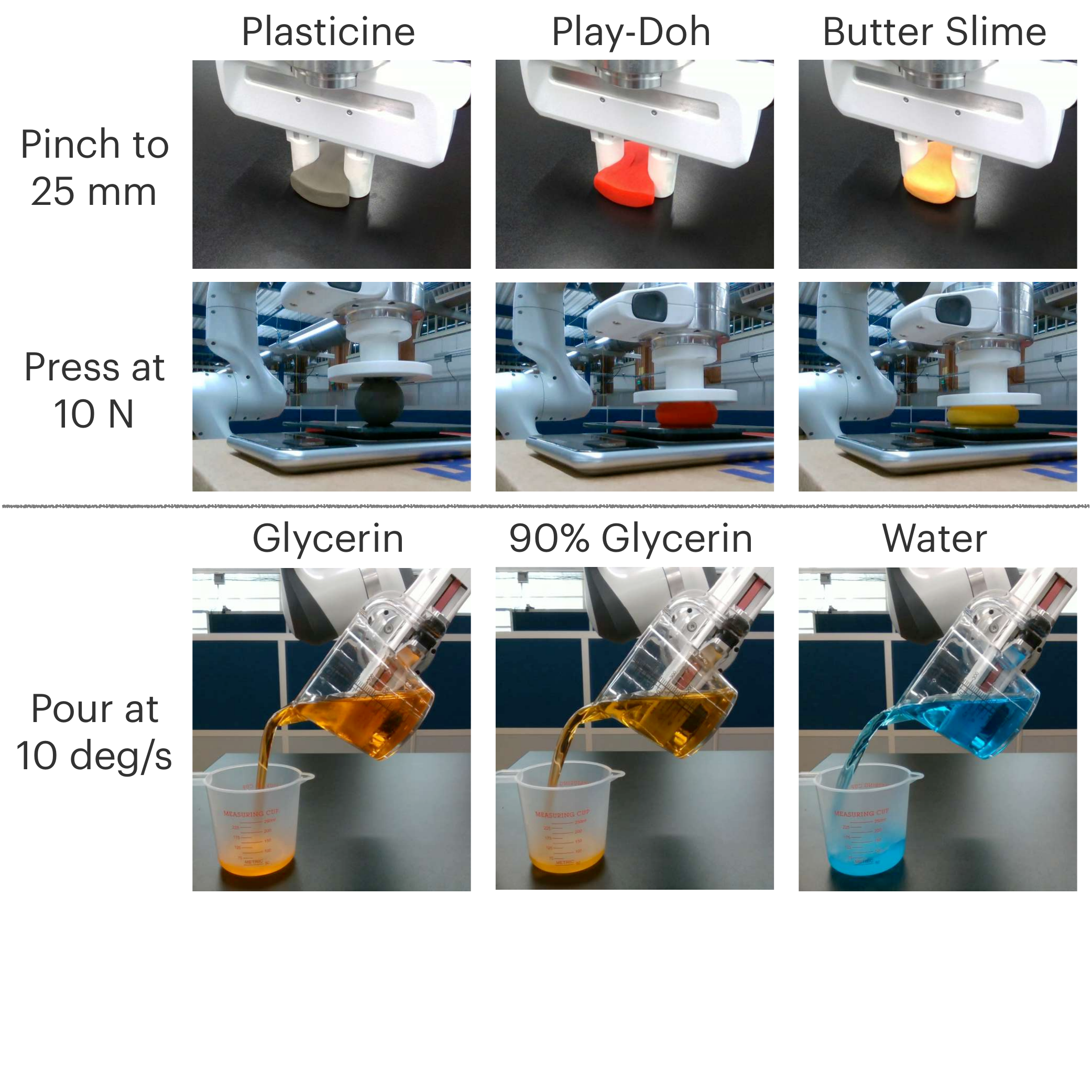}
    \vspace{-23mm}
    \caption{The dynamics of deformable materials and fluids are often hard to observe from vision alone. Active physical interaction is required to estimate their latent properties and enable robust manipulation policies. \textbf{(Top)} Starting from the same initial shape, applying the identical robot action such as a 25 mm pinch or a 10 N press results in different deformations dictated by the stiffness and yield behavior. \textbf{(Bottom)} Similarly, applying the same pouring action at 10 degrees per second results in distinct flow behaviors governed by the varying viscosities of the fluids.}
    \label{fig:teaser}
\end{figure}

An unfamiliar deformable material presents a robot with unknown responses to force and motion.
These differences directly affect manipulation: two pieces of dough can deform differently under the same press, and two liquids can flow at different rates under the same pouring motion (\Cref{fig:teaser}).
A robot can infer material behavior from observed motion and, when needed, interaction forces. The recovered model must then predict how the material will respond to actions that have not yet been executed.
A constitutive law provides such a model by relating deformation and deformation rate to internal stress, allowing prediction under new actions.

Existing approaches either avoid explicit material identification, require costly iterative optimization, or are not designed for robot interaction data.
Learned dynamics models predict motion from prior interaction data, but leave constitutive behavior implicit in the learned representation~\cite{hansen2024td,khorrambakht2025worldplanner,thorpe2026physically}.
Differentiable system-identification methods recover explicit material parameters, but require repeated simulation of the observed interaction and differentiation through the resulting trajectory~\cite{ma2023learning,li2023pac,yang2025differentiable}.
Weak-form identification avoids repeated rollout matching by enforcing momentum balance directly on measured motion and loads~\cite{pierron2012virtual,flaschel2021unsupervised,flaschel2022discovering}.
However, these methods have largely been developed for controlled mechanical tests on prepared specimens, rather than the partial, time-varying observations available during robot interaction.

FORM reduces material identification from robot interaction to linear least-squares solves whose solutions are used directly by the robot's simulator.
The robot tracks points on the material to reconstruct deformation during a probe and, when needed, measures contact force.
We express the stress as a weighted sum of candidate constitutive responses. Enforcing momentum balance on the observed motion then gives a linear system whose unknowns are the material coefficients.
We assemble this system in the same material point method (MPM) discretization used for forward prediction, so the recovered coefficients can be used directly for planning without refitting or conversion.

The experiments test whether a material model identified from one interaction remains useful under a different action or geometry.
We first evaluate identification accuracy and solve time across elastic, granular, elastoplastic, and fluid materials under new motions, initial conditions, and geometries.
We then use the identified models to plan rod insertion, golf putting with an elastic club, elastoplastic shaping, and target-volume pouring, with deployment motions different from the identification probes.
In every case, the identified model is held fixed during deployment, with no refitting from execution measurements.
Across these experiments, elastic properties are recovered within 3.4\%, elastoplastic properties within 2\%, hardware shaping reaches 72.4--77.8\% intersection over union (IoU), and pouring reaches at most 3.8~mL mean error.

Our contributions are:
\begin{enumerate*}[label=(\arabic*)]
\item \textbf{Direct constitutive identification from robot interaction:} which recovers material laws from tracked motion and, when required, contact forces. Each identification stage requires only a single linear solve, assembled in the same MPM discretization used for prediction.

\item \textbf{Transfer from probing to manipulation:}
we show that models identified from a single interaction can be reused without refitting to predict and plan under new actions and geometries across elastic, elastoplastic, and fluid materials.
\end{enumerate*}

Videos are available on the project page, \url{https://form-robots.github.io/}, and code at \url{https://github.com/form-robots/FORM}.

\section{Related Work}
\label{sec:related}

Controlling deformable objects is challenging due to their high-dimensional configurations, nonlinear dynamics, limited observability, and strong underactuation \cite{hu20193,arriola2020modeling,zhu2022challenges,ying2024obstacle}. Model-based control therefore requires predictive models that are both accurate and computationally tractable. RoboCraft and RoboCook learn particle-based dynamics with graph neural networks~\cite{shi2024robocraft,shi2023robocook}, while DiffSkill and DiffVL use differentiable MPM simulators~\cite{lin2022diffskill,huang2023diffvl}. For liquids, SPNets embeds differentiable particle-based fluid dynamics in a neural network, while adaptable pouring estimates parameters of a simplified fluid simulator from observed behavior~\cite{schenck2018spnets,lopez2017adaptable}.
Long-horizon model-based planning methods often compose skills or optimize reduced action parameters or subgoals~\cite{lin2022planning,huang2023diffvl,you2025make,pmlr-v305-yamada25b}.
Although these methods achieve strong results within their respective regimes, changes in material properties often require the predictive model or simulator to be retrained, recalibrated, or otherwise adapted to reflect the new material behavior.

Differentiating through learned or physics-based simulators such as MPM and SPH remains the dominant approach to material identification. Direct optimization can accurately identify material parameters~\cite{ma2022risp,chen2026empm}, but requires specifying the constitutive law family. NCLaw instead learns a neural constitutive model by matching trajectories through differentiable MPM~\cite{ma2023learning}. PAC-NeRF jointly reconstructs geometry and physical parameters from multiview video~\cite{li2023pac}, while DPSI estimates elastic, plastic, and frictional parameters from a single robot interaction~\cite{yang2025differentiable}.
Material-conditioned dynamics models provide another route by training shared graph-based simulators across material variations and adapting a compact set of conditioning variables to new materials~\cite{zhang2024adaptigraph,manoharan2025parameter}. 
These approaches require optimizing highly nonconvex objectives through repeated simulation and backpropagation through trajectories, which can be computationally expensive and unstable over long horizons.

Weak-form constitutive identification offers an alternative to simulation-based system identification. The virtual-fields method recovers material parameters by enforcing equilibrium against selected virtual displacement fields~\cite{pierron2012virtual,mei2019improving}. The EUCLID family extends this principle to discover interpretable constitutive laws from full-field kinematics and global reaction forces, using sparse regression over candidate libraries for hyperelasticity, plasticity, and generalized standard materials~\cite{flaschel2021unsupervised,flaschel2022discovering,flaschel2023automated}. These methods operate directly on measured motion and loads, avoiding repeated forward simulation and backpropagation through trajectories. Candidate libraries reduce assumptions about the constitutive law by spanning multiple material models.
However, existing methods generally require full-field measurements that are difficult to obtain in robotic settings, where stress and internal pressure may be inaccessible. They are also formulated primarily for quasi-static solid mechanics rather than time-evolving particle dynamics, preventing direct application to freely deforming granular materials and fluids.

\section{Identifying the Material from Motion}
\label{sec:method}
\label{sec:laws}

\begin{figure*}[t]
\centering
\includegraphics[keepaspectratio,width=\textwidth]{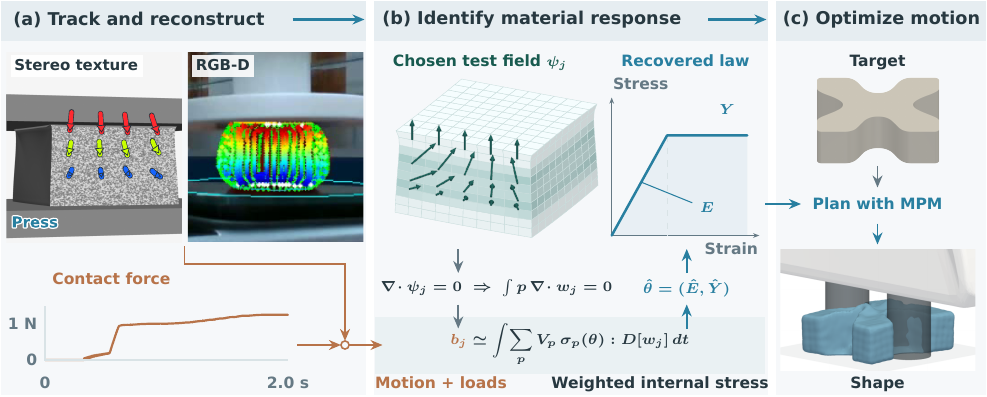}
\vspace{-7mm}
\caption{FORM: material identification and manipulation. (a) Synthetic stereo tracks and motion reconstructed from real RGB-D observations. Colors indicate image displacement in stereo and initial height in RGB-D. The force trace is from the synthetic press. (b) Divergence-free test fields remove pressure. Green arrows denote test vectors, not motion. The weak balance identifies stiffness $E$ and yield threshold $Y$. The stress–strain curve is schematic. (c) The recovered model and target are used to plan the robot motion in MPM, illustrated here in the simulated shaping task.}
\label{fig:method-overview}
\end{figure*}

We introduce FORM (From Observed Response to Material laws), a weak-form method that identifies a material's constitutive law from an observed robot interaction and uses it to predict the material's response to new actions. During identification, we hold the observed motion fixed and express stress as a linear combination of candidate responses. Momentum balance then gives a linear inverse problem in the coefficients without repeated trial simulations, even when deformation is large. During prediction, the simulator holds those coefficients fixed and evaluates the law as the material state evolves. The method outline is shown in \Cref{fig:method-overview}.

Surface motion can be recovered from visual observations, for example through stereo texture tracking or RGB-D shape reconstruction. Reconstructing motion inside the material additionally requires geometric and kinematic assumptions.
The reconstructed motion gives position histories $x_p(t)$ and velocities $v_p(t)=dx_p(t)/dt$. Spatial derivatives of this motion define the deformation gradient $F$ and velocity gradient $L$, relating local separations and velocity differences through
\begin{equation}
\label{eq:kinematics}
\Delta x(t)\approx F(t)\Delta X,\qquad
\Delta v(t)\approx L(t)\Delta x(t),
\end{equation}
where $\Delta x$ is the neighbors' current separation, $\Delta v$ their velocity difference, and $\Delta X$ their separation in the reference configuration. Elastic laws use $F$ to determine stress from accumulated deformation. Viscous laws use the stretching and shearing rate $D=(L+L^\top)/2$, where $L=\nabla v$ is the velocity gradient. The stress $\sigma$ is not observed, but it must produce the observed motion through momentum balance,
\begin{equation}
\label{eq:momentum}
\nabla\cdot\sigma+\rho g=\rho a,\qquad \sigma=-pI+\tau,
\end{equation}
where $\rho$ is density, $g$ is gravity, and $a$ denotes acceleration.
The isotropic part $-pI$ resists volume change, and the deviatoric part $\tau$ resists shape change.

\subsection{Removing pressure to identify viscosity}
We begin with a simple example to illustrate the key ideas behind our method.
For a Newtonian fluid with constant viscosity $\eta$ and negligible bulk viscosity, the full  constitutive law is
\begin{equation}
\label{eq:newtonian}
\sigma=-pI+\tau,\qquad
\tau=2\eta\,\mathrm{dev}\,D,
\end{equation}
where $\mathrm{dev}\,D=D-\tfrac13\mathrm{tr}(D)I$.
Similar to the stress split in~\eqref{eq:momentum}, we split $D$ into the rate of volume change, its trace, and the rate of shape change $\mathrm{dev}\,D$, which viscosity resists.

Motion determines $\mathrm{dev}\,D$, but the balance still couples the viscosity $\eta$ we seek to an unobserved pressure field $p(x,t)$.
We determine viscosity without reconstructing pressure by choosing weighted balances that remove its contribution.

We weight momentum balance with a smooth vector field $\psi_j$ that vanishes near the boundary of the current body $\Omega$.
Integration by parts removes the boundary term and gives the weak formulation, with derivatives on the weight rather than on stress,
\begin{equation} \label{eq:weak}
\begin{aligned}
\eta\underbrace{\int_\Omega 2\,\mathrm{dev}\,D:D[\psi_j]\,d\Omega}_{A_j}
&-\int_\Omega p\,\nabla\cdot\psi_j\,d\Omega\\
&=\underbrace{\int_\Omega\rho(g-a)\cdot\psi_j\,d\Omega}_{b_j},
\end{aligned}
\end{equation}
where $D[\psi_j]=(\nabla\psi_j+\nabla\psi_j\T)/2$.
Pressure now multiplies the divergence of a test function we choose.
Choosing $\nabla\cdot\psi_j=0$ everywhere makes the pressure integral zero for every pressure field, leaving
\begin{equation}
\label{eq:pressure-cancellation}
A_j\eta=b_j, \quad\text{for}\ j = 1, 2, \dots, n.
\end{equation}
Each weight function gives an equation for the same viscosity, which can be determined if at least one $A_j$ is nonzero.

Choosing $\psi_j=\nabla\times\boldsymbol{\xi}_j$, with a smooth vector potential $\boldsymbol{\xi}_j$ supported inside the body, gives $\nabla\cdot\psi_j=0$ identically.
The zero-divergence condition applies to the weight, and the material may still compress, so pressure cancellation also holds for weakly compressible motion.

\subsection{The same spatial basis for observations and MPM}

A key advantage of our weak formulation is that we can choose weight functions to match the spatial stress assembly of the forward MPM solver.
Its quadratic B-splines $N_i$ define a vector weight in each coordinate direction $e_d$:
\begin{equation}
\label{eq:shared-basis}
\psi_{i,d}(x)=N_i(x)e_d.
\end{equation}
For symmetric stress, this weighted balance recovers the MPM internal-force assembly under particle quadrature,
\begin{equation}
\label{eq:shared-force}
\int_\Omega\sigma:D[\psi_{i,d}]\,d\Omega
\approx\sum_p V_p e_d\T\sigma_p\nabla N_i(x_p)=-e_d\T f_i^{\mathrm{int}}.
\end{equation}
Here particle $p$ has volume $V_p$, and $f_i^{\mathrm{int}}$ is the MPM internal force at grid node $i$. During identification, the B-splines and their gradients are evaluated at positions reconstructed from the observed motion; during prediction, they are evaluated at simulated particle positions.

\subsection{Representing a range of constitutive laws}

\begin{table}[t]
\caption{Constitutive models and parameters, with $\sigma=-pI+\tau$. Stresses are Cauchy unless marked $\mathrm{K}$ for Kirchhoff.}
\label{tab:laws}
\centering
\scriptsize
\setlength{\tabcolsep}{2.5pt}
\renewcommand{\arraystretch}{1.15}

\renewcommand{\tabularxcolumn}[1]{m{#1}}

\begin{tabularx}{\columnwidth}{
    @{}
    >{\raggedright\arraybackslash}m{0.20\columnwidth}
    >{\raggedright\arraybackslash}X
    >{\raggedright\arraybackslash}m{0.19\columnwidth}
    @{}
}
\toprule
Law & Stress response & Parameters \\
\midrule

\multicolumn{3}{@{}l}{\textbf{Elastic and hyperelastic}, with $p=-\kappa(J-1)$ for the $\tau$ rows} \\
\cmidrule{1-3}

neo-Hookean &
$\displaystyle
\tau=\frac{2C_{10}}{J}\,\mathrm{dev}\,\bar b$ &
$C_{10},\kappa$ \\[3pt]

Mooney--Rivlin &
$\displaystyle
\tau=\frac{2}{J}\,\mathrm{dev}\!\left[C_{10}\bar b
+C_{01}(\bar I_1\bar b-\bar b^2)\right]$ &
$C_{10},C_{01},\kappa$ \\[3pt]

Yeoh &
$\displaystyle
\tau=\frac{2}{J}\Big[\sum_{i=1}^{3}iC_i(\bar I_1-3)^{i-1}\Big]
\mathrm{dev}\,\bar b$ &
$C_1,C_2,C_3,\kappa$ \\[3pt]

Gent &
$\displaystyle
\tau=\frac{\mu}{J\big[1-(\bar I_1-3)/J_m\big]}\,
\mathrm{dev}\,\bar b$ &
$\mu,J_m,\kappa$ \\[3pt]

fixed-corotated &
$\displaystyle
\sigma=\frac{2\mu}{J}(F-R)F^{\mathsf T}
+\lambda(J-1)I$ &
$\mu,\lambda$ \\

\addlinespace[2pt]
\midrule
\multicolumn{3}{@{}l}{\textbf{Inelastic and flow}} \\
\cmidrule{1-3}

von Mises (J2) &
$\begin{gathered}
\tau_i^{\rm K}=2G\varepsilon_{e,i}
+\lambda\,\mathrm{tr}\,\varepsilon_e,\;
\|\mathrm{dev}\,\tau^{\rm K}\|_F\le\tau_y
\end{gathered}$ &
$G,\lambda,\tau_y$ \\[3pt]

granular $\mu(I)$ (incompressible flow) &
$\begin{gathered}
\tau=\mu(I)p\frac{2D}{\dot\gamma_\epsilon},\quad
\mu(I)=\mu_s+\frac{\Delta\mu\,I}{I+I_0}
\end{gathered}$ &
$\mu_s,\Delta\mu,I_0$ \\[3pt]

Newtonian / regularized Bingham &
$\displaystyle
\tau=2\eta\,\mathrm{dev}\,D
+2\tau_y\frac{\mathrm{dev}\,D}
{\dot\gamma_\epsilon}$ &
$\eta,\tau_y$ \\

\bottomrule
\end{tabularx}

\vspace{2pt}
{\footnotesize\raggedright
$F=RU$ is the polar decomposition, $J=\det F$, $\bar b=J^{-2/3}FF^{\mathsf T}$, and $\bar I_1=\mathrm{tr}\,\bar b$. The quantities $\varepsilon_{e,i}$ are the principal elastic logarithmic strains. The regularized shear rate is $\dot\gamma_\epsilon=\sqrt{2\,\mathrm{dev}\,D:\mathrm{dev}\,D+\epsilon^2}$, with regularization parameter $\epsilon>0$. In the granular law, $I$ is the inertial number.}
\end{table}

While Newtonian viscosity requires only one coefficient $\eta$, we want to represent a range of constitutive laws. Following EUCLID~\cite{flaschel2023automated}, 
we therefore replace the shear response $2\eta\,\mathrm{dev}\,D$ with a linear combination of functions that can depend nonlinearly on the material state,
\begin{equation}
\label{eq:dict}
\tau(q;\theta)=\sum_{k=1}^{K}\theta_kT_k(q).
\end{equation}
Here $q$ contains the deformation or rate quantities reconstructed from motion or supplied from measurements and can differ by material.
The Newtonian fluid is the case $K=1$, with $T_1=2\,\mathrm{dev}\,D$ and $\theta_1=\eta$.
The functions $T_k$ describe possible stress responses, and their coefficients $\theta_k$ specify the material.
Linearity is required only in the coefficients, and each $T_k$ may depend nonlinearly on deformation.
Candidates for such basis functions come from existing constitutive models. 
Table~\ref{tab:laws} lists the analytic laws and their physical parameters.
Laws linear in those parameters supply basis functions directly, as viscosity and yield stress do for regularized Bingham flow.

For materials whose constitutive law is unknown or difficult to determine a priori, a learned set of neural basis functions can be substituted as a data-driven alternative. In particular, we adopt the function encoder method from~\cite{ingebrand2025basis}, where the bases are trained offline from generated constitutive laws and can span a family of plausible constitutive responses for each material type. For instance, learned hyperelastic bases cover the neo-Hookean, Yeoh, and Gent families and apply to any material with hyperelastic behavior.
Function encoder bases retain the linear-in-coefficients property, so online identification remains one least-squares solve.

We note that the spatial basis $N_i$ determines how stress enters each balance, and the constitutive basis $T_k$ determines which responses that balance can identify.
Full-stress identification extends~\eqref{eq:dict} from the deviatoric stress to the full stress, $\sigma=-pI+\sum_k\theta_kT_k$.
The sum now includes volumetric responses such as $(J-1)I$, and the first term collects any contribution known before the fit, a supplied pressure $-pI$ in our experiments and zero otherwise.
The state $q$ and this known contribution must be available independently of the fitted coefficients.

Plastic laws require elastic state $F_e$, and pressure-dependent friction requires pressure even when the weights remove isotropic stress and thus require specific treatment. 
For sand, the deviatoric stress depends directly on pressure, meaning divergence-free test fields cannot fully eliminate it from the momentum balance.
We instead compute this pressure from the observed volume change using the equation of state $p=-\kappa\log(J_e)/J_e$, supplying it as fixed data so the linear solve gains no new unknowns.

Finally, estimating acceleration from tracked motion is imprecise at high deformation rates.
We therefore use separable weights $w_j(x,t)=\chi_j(t)\psi_j(x)$, where $\chi_j$ vanishes at both window endpoints.
The separability preserves pressure cancellation and integration by parts replaces acceleration with the derivative of the known weight along each track,
$\dot w_{jp}=\partial_t w_j(x_p,t)+(\nabla w_j(x_p,t))v_p$.
The resulting coefficient system is $A\theta\simeq b$, with
\begin{equation}
\label{eq:system}
\begin{aligned}
A_{jk}&=\int\sum_p V_pT_k(q_p):D[w_j]\,dt,\\
b_j&=\int\sum_p\big[m_p(g\cdot w_j+v_p\cdot\dot w_{jp}) +V_p\,p\,\nabla\!\cdot\!w_j\big]\,dt+c_j.
\end{aligned}
\end{equation}
We evaluate the integrands at the particle positions $x_p(t)$ and integrate over the time window.
The particle masses $m_p$, current volumes $V_p$, and pressure $p$ are known inputs.
We set the test functions to zero at fixed boundaries, which eliminates the unknown reaction forces from the balance.
At boundaries with prescribed velocity, such as the pressing plate, the test function equals that velocity, and we compute $c_j=\int\chi_j\,f_c\cdot v_c\,dt$ from the measured force.

Each test function gives one equation, and each stress basis function $T_k$ gives one column of $A$.
We combine these equations across samples and time windows and fit one set of material coefficients by least squares,
\begin{equation}
\label{eq:solve}
\hat\theta=\underset{\theta}{\arg\min}\;\|A\theta-b\|_2^2.
\end{equation}
For liquid flowing from a source cup into a receiver, we apply the same time-integrated weak-form approach to a reduced model of the momentum balance \eqref{eq:momentum} at the cup lip. We infer the liquid head from receiver-volume measurements, recorded cup motion, known geometry, and initial fill, assuming a horizontal free surface. Integrating the reduced balance over time gives a linear fit for inverse viscosity from receiver-volume increments, without reconstructing interior motion.

\subsection{Numerical Validation and Comparison}

\begin{table*}[t]
\centering
\caption{Reconstruction results for FORM and baselines. Variants use learned bases selected by material type (\textbf{Ours (Fn Enc.)}), full information (\textbf{Ours (Full)}), observations without stress (\textbf{Ours (No stress)}), or position-only observations (\textbf{Ours (Pos. only)}). Bold marks a lowest-loss non-oracle result and its solve time.}
\label{tab:generalization}

\resizebox{\textwidth}{!}{%
\begin{tabular}{l|cc|cc|cc|cc}
\toprule

& \multicolumn{2}{c|}{\textsc{Jell-O}}
& \multicolumn{2}{c|}{\textsc{Sand}}
& \multicolumn{2}{c|}{\textsc{Plasticine}}
& \multicolumn{2}{c}{\textsc{Water}} \\
\textbf{Method}
& \textbf{Mean Loss} & \textbf{Solve Time (s)}
& \textbf{Mean Loss} & \textbf{Solve Time (s)}
& \textbf{Mean Loss} & \textbf{Solve Time (s)}
& \textbf{Mean Loss} & \textbf{Solve Time (s)}
\\

\midrule

\textbf{NCLaw}
& 5.2e-4 & 1513.2
& 1.6e-4 & 1305.6
& 1.2e-4 & 1300.2
& 5.7e-4 & 1306.8
\\

\textbf{Diff. Sys-Id}
& 1.2e-8 & 715.8
& 6.3e-13 & 1132.8
& 1.6e-10 & 1028.4
& 4.0e-9 & 556.2
\\

\textbf{Ours (Fn Enc.)}
& 3.5e-4 & 5.2
& 3.3e-8 & 0.1
& 2.9e-6 & 5.3
& 2.3e-6 & 4.8
\\

\textbf{Ours (Full)}
& 5.1e-6 & 4.4
& \textbf{1.4e-8} & \textbf{2.4}
& \textbf{5.2e-7} & \textbf{4.7}
& 2.3e-6 & 2.2
\\

\textbf{Ours (No stress)}
& \textbf{5.1e-6} & \textbf{3.5}
& 2.0e-8 & 6.0
& 5.2e-7 & 6.5
& \textbf{2.3e-6} & \textbf{1.6}
\\

\textbf{Ours (Pos. only)}
& 1.3e-5 & 3.5
& 1.0e-5 & 12.8
& 4.6e-6 & 21.8
& 1.0e-2 & 4.6
\\

\bottomrule
\end{tabular}%
}

\end{table*}

We compare FORM against numerical identification baselines in~\Cref{tab:generalization}.
We mainly use the dataset and follow the setup in~\cite{ma2023learning}, and consider four representative material types: Water, Sand, Jell-O, and Plasticine, each having different internal constitutive models.
We compare against NCLaw~\cite{ma2023learning}, which learns constitutive models from visual trajectories, and an oracle system identification scheme that differentiates through the MPM trajectories assuming exact constitutive models are known. One trajectory per material type is used for inference across all testing scenarios.
For plasticine and sand, positions alone do not determine the elastic state, so \textbf{Ours (Pos. only)} scans forward rollouts over the plastic parameter with the elastic moduli fixed at their true values.
We evaluate generalization under three conditions for each example: extended time, unseen initial conditions, and different geometry. Reported accuracies and solve times are averaged across scenarios. As shown in~\Cref{tab:generalization}, FORM generally achieves competitive approximation accuracy while incurring only a fraction of the solve time required by the baselines. In cases where exact constitutive laws are unknown and only general material types are available, pretrained function encoder bases can be used under the same framework, sacrificing negligible performance while maintaining the fast inference speed not possible with existing methods.

\section{Material-Conditioned Robot Motion Optimization}
\label{sec:planning}
We optimize robot motions conditioned on the identified material parameters $\hat\theta$. For each task, we prescribe an action sequence and optimize a low-dimensional set of motion parameters $u\in\mathcal{U}$ where $\mathcal U$ is the admissible set. These parameters define the robot motion while keeping the search space small enough for efficient evaluation of candidate motions through simulation.

For each candidate $u$, MPM predicts the material response from the initial state $x_0$ with $\hat\theta$ held fixed. 
We represent the tools as meshes and resolve material-tool interactions by imposing the corresponding boundary conditions during MPM simulation. The simulation time step is selected to satisfy the constraints defined by these boundaries.

\begin{figure*}[b]
\centering
\includegraphics[width=\textwidth]{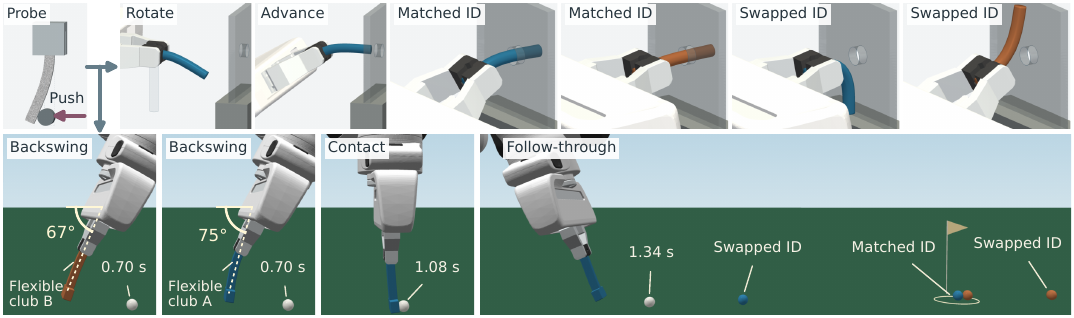}
\vspace{-7mm}
\caption{Stiffness from a common bending probe transfers to rod insertion (top) and elastic-club putting (bottom).
Blue and orange denote materials A and B.
The pale outline marks the initial rod pose.
Putting shows B and A at the end of the backswing, then A at contact and follow-through.
Arcs measure projected grip-to-head inclination relative to horizontal; white balls and timestamps show recorded motion states.
Colored balls overlay final positions from four separate executions, with color denoting club material.
The circle marks the stopping target.}
\label{fig:bending-insertion}
\end{figure*}

We denote the predicted trajectory by $\mathcal{S}_{\hat\theta}(x_0,u)$ and seek
\begin{equation}
\label{eq:motion-optimization}
u^\star(\hat\theta)\in\underset{u\in\mathcal{U}}{\arg\min}\;
\mathcal{J}\!\left(\mathcal{S}_{\hat\theta}(x_0,u),\mathcal{G}\right),
\end{equation}
where $\mathcal{G}$ is the task goal and $\mathcal{J}$ measures task error and any constraint penalties. For example, in simulated shaping, $\mathcal{G}$ is the target surface and $\mathcal{J}$ is the symmetric, area-weighted mean closest-point distance between the released specimen surface and the target surface. In pouring, $\mathcal{G}$ is the target receiver volume and $\mathcal{J}$ is the absolute volume error.

We search over the admissible motion parameters by repeatedly simulating candidate actions and evaluating their cost $\mathcal{J}$. The search terminates when its stopping criterion is met or the evaluation budget is exhausted, and we return the best motion evaluated.
We refer to~\Cref{sec:results} for task-specific planning details.

Finally, we execute the selected motion open loop in simulation and on hardware. Measurements collected during execution, including force, shape, and poured volume, are not used to update the material model or replan the motion.

\section{Results}
\label{sec:results}

In simulation and on a real Franka Panda robotic arm, we test whether a material law identified from a single probing interaction can guide motion planning and manipulation involving deformable materials.
To demonstrate the importance of correct identification, we compare task execution using the correct material identification (\textit{matched ID}) against execution using an incorrect material identification (\textit{swapped ID}).

\subsection{Elastic Material Manipulation}

We tackle two simulated tasks with a Franka Panda arm: 1) deformable peg insertion (Fig.~\ref{fig:bending-insertion}, top row), and 2) golf putting where an elastic club strikes a ball to stop inside a circular target (Fig.~\ref{fig:bending-insertion}, bottom row). For both tasks, we plan the robot's motion using a material model identified by bending a rectangular strip of the same material. During execution, we prescribe the robot's joint motion and keep the grasped end of the peg or club fixed to the gripper. Both manipulation tasks are evaluated with two homogeneous elastic materials, A and B, with Young's moduli of $80$ and $240$~kPa, respectively. 

We track surface texture on the strip in synthetic stereo images to recover its surface motion. Interior reconstruction assumes uniform in-plane displacement through the strip's thickness, with transverse stretch determined by a plane-stress relation. We estimate Young's modulus from the reconstructed motion and pusher force using the weak momentum balance in Sec.~\ref{sec:method}. Geometry, camera calibration, density and Poisson's ratio $\nu=0.45$ are known. No simulator particle states enter the fit. For A and B, the estimates are $80.61$ and $248.11$~kPa, with errors of $0.76\%$ and $3.38\%$, respectively.
\begin{figure*}[t]
\centering
\includegraphics[width=\textwidth,trim=0 0 0 0,clip]{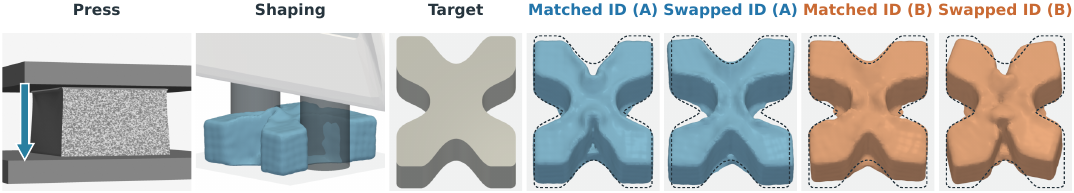}
\vspace{-7mm}
\caption{Material identification and X shaping in simulation.
Left: A's identification press and final planned pinch. The arrow indicates plate motion; translucent cylinders show the simulated fingers, with the hand illustrating their pose.
Right: target shape and matched and swapped results for A (blue) and B (orange), all at a common scale. Results are shown $1$~s after finger withdrawal; dashed outlines mark the target.}
\label{fig:shaping}
\end{figure*}

\begin{figure}[!b]
\centering
\includegraphics[width=\columnwidth,trim=0 1mm 0 1mm,clip]{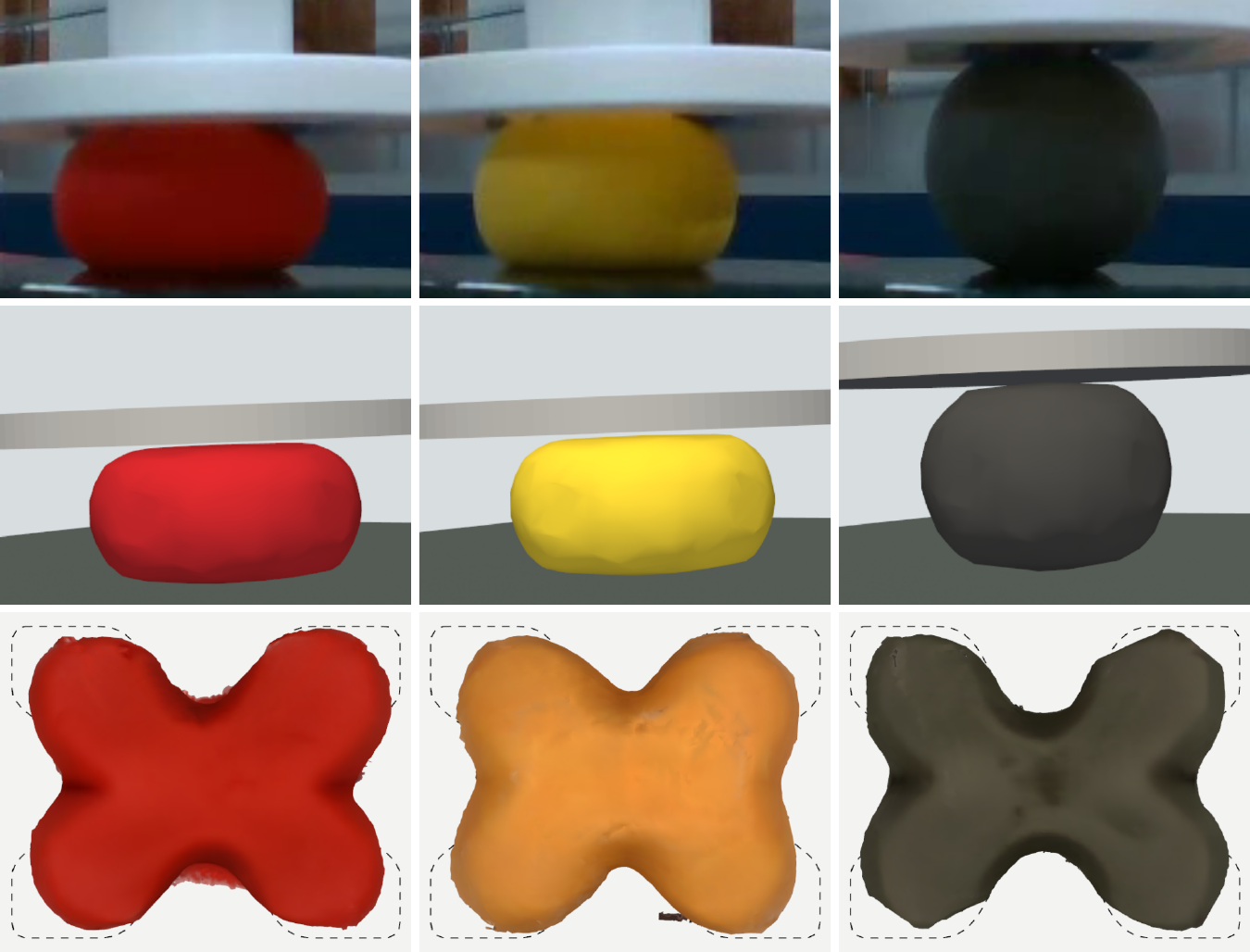}
\vspace{-7mm}
\caption{Hardware pressing and shaping of red Play-Doh, yellow butter slime, and gray plasticine (left to right). Top: recorded pressing. Middle: force-driven MPM predictions at matching times, using models identified from separate presses. Simulation views are horizontally mirrored for display. Bottom: reconstructed shapes after four open-loop pinches, rigidly aligned to the dashed target outlines without scaling.}
\label{fig:hardware-shaping}
\end{figure}
Peg insertion requires the robot to thread a deformable rod through a circular opening in a wall. The rod is $60$~mm long and $8$~mm in diameter, and the hole is $12$~mm in diameter. We optimize gripper height and tilt to cross the wall while penalizing misalignment and insufficient clearance.
Both matched plans insert without wall contact. Over the final $0.5$~s hold, the RMS distance from the rod-tip centroid to its target is $0.355$~mm for A and $1.042$~mm for B. Both swapped plans collide with the wall and fail to insert.

In the golf task, the goal is to stop a rigid $16$~mm diameter ball entirely inside a circular target of radius $35$~mm, centered $612$~mm from the ball's initial position. With the backswing fixed, we optimize forward-stroke duration and aim angle to minimize the final distance between the ball and target center.
We assume known ball properties, floor friction and rolling resistance, with frictionless club-ball contact. Both matched plans stop the ball entirely inside the target, with center errors of $2.36$ and $19.06$~mm, respectively.
The swapped plans miss the target, with center errors of $350.47$ and $158.67$~mm.

\subsection{Elastoplastic Material Shaping}
Elastoplastic shaping is highly material-dependent: stiffness and yield behavior determine how much permanent deformation each pinch produces, so using the wrong material model can cause the robot to over- or under-deform the object and miss the target shape.
To evaluate our method on elastoplastic materials, we design a shaping task where the robot transforms a square block into a rounded X-shaped prism (Fig.~\ref{fig:shaping}). This transformation is achieved through a series of alternating horizontal and vertical pinches using cylindrical gripper fingers. To identify the underlying material model, we first compress a smaller specimen with a flat probe while it rests on a rigid support. We perform the experiment in both simulation and on the real robot, evaluating our method across two simulated and three physical materials.

\textbf{Simulated Material Identification.} Let A and B denote two homogeneous materials that follow Hencky elasticity with perfect von Mises plasticity.
Both share the Young's modulus, $E=80$~kPa, but have different yield thresholds, $Y=1$ and $10$~kPa, respectively.
For identification, a flat plate compresses a $30\times30\times25$~mm specimen without unloading.
We recover surface motion by tracking texture on the specimen in synthetic stereo images, which we extend into the interior using a field constrained by the specimen's square symmetry.
The reconstructed motion and normal plate force enter the weak momentum balance using analytic divergence-free test fields.
By assuming an early loading interval is elastic and a later interval is fully yielded, we estimate $E$ and $Y$ through two scalar least-squares fits.
The estimates are $(E,Y)=(80.795,1.015)$~kPa for A and $(81.396,10.184)$~kPa for B, with all errors below $2\%$.

\textbf{Hardware Material Identification.} To identify the physical parameters of red Play-Doh, yellow butter slime, and gray plasticine, we apply the same weak-form method to real-world data. We recover the models beforehand by applying a downward, flat-surface press to a spherical specimen with a 45~mm diameter. We track the material using a pair of RealSense D435 RGB-D cameras to capture the silhouette of the pancake shape as the sphere is compressed. For this reconstruction, we assume constant volume and that particles touching the top and bottom surfaces become stationary.

\textbf{Planning.} We use the identified models to plan pinch sequences on fresh $60\times60\times25$~mm blocks, targeting a common X-shaped footprint. The gripper, inspired by RoboCraft~\cite{shi2024robocraft}, features two $28$~mm diameter cylindrical fingers that close from opposite sides, rotating $90^\circ$ between each pinch.
Simulation uses six pinches and optimizes the final opening of each pinch and a shared lateral offset for the first four. Hardware experiments uses four pinches only.
In both domains, the planners use the specific material models recovered during the identification phase to refine the sequence and minimize final shape error.

\begin{figure*}[t]
\centering
\includegraphics[width=\textwidth,trim=0 0 0 0,clip]{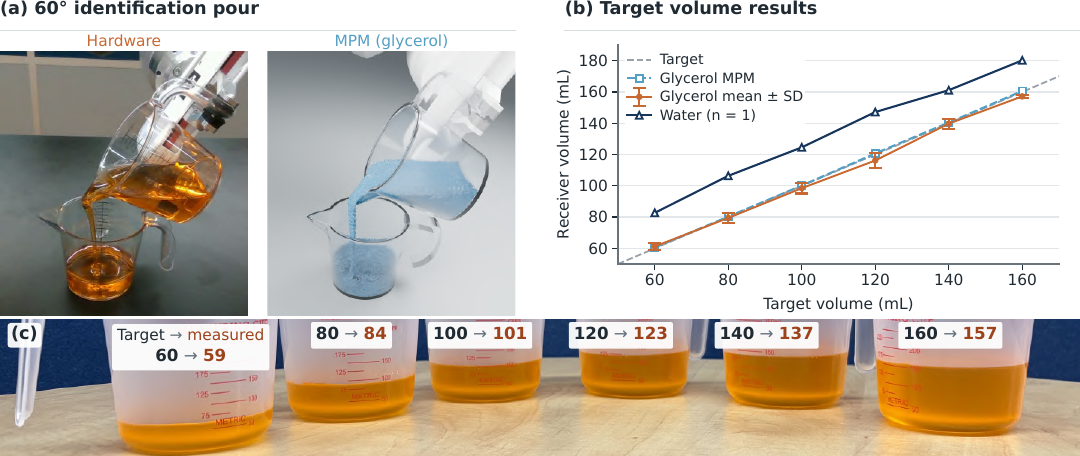}
\vspace{-7mm}
\caption{Glycerol pouring with an identified model.
(a) Hardware and MPM at matching times during the $60^\circ$ identification pour; this recording supplies both viscosity identification and effective contact calibration. Simulated liquid is colored blue for visibility.
(b) Glycerol: frozen MPM predictions and hardware mean $\pm$ sample standard deviation over five trials per target. Water: one hardware trial per target, without error bars.
(c) Repeat 2: target $\rightarrow$ measured receiver volume (mL), with readings rounded to the nearest mL.}
\label{fig:pouring}
\end{figure*}

\textbf{Evaluation.} In simulation, full state access allows us to measure surface error as the mean closest-point distance between the specimen and target surfaces $1$~s after finger withdrawal. Matched plans significantly outperform swapped plans, yielding errors of $0.883$ and $1.256$~mm for materials A and B, respectively, compared to $1.226$ and $1.739$~mm when the models are swapped (Fig.~\ref{fig:shaping}). On hardware, the lack of ground-truth internal particle tracking prevents us from precisely evaluating volumetric deformation (Fig.~\ref{fig:hardware-shaping}). We therefore rely on reconstructed surface scans and evaluate the physical shapes using the intersection over union (IoU) of their $XY$ footprint against the target.
The resulting IoUs are $75.7\%$, $72.4\%$, and $77.8\%$ for the Play-Doh, butter slime, and plasticine, respectively.

\subsection{Pouring a Target Volume}
We evaluate fluid manipulation with a target-volume pouring task.
The real robot must transfer a requested volume of glycerol from a source cup, initially filled to its $300$~mL graduation, to a receiver cup (Fig.~\ref{fig:pouring}). Because the exact viscosity is unknown beforehand, an exploratory pour is necessary to estimate this latent property and enable precise volume targeting.

\textbf{Material Identification.} We estimate the fluid's viscosity from a single $60^\circ$ exploratory pour.
We determine the receiver volume over time by segmenting the liquid by color and tracking its free-surface height. Separately, we compute the source cup's kinematics directly from the recorded joint trajectories of the robot arm.
We apply the weak-form approach to a reduced model of flow at the source cup's spout. Integrating the momentum balance over time gives an effective Newtonian viscosity of $3.44$~Pa\,s. For the subsequent MPM simulation, we also use this same exploratory pour's final received volume to calibrate an effective wall-contact coefficient of $0.272$, keeping the estimated viscosity fixed.

\textbf{Planning.} Using the identified fluid model, we optimize a single control parameter, the source cup's final tilt angle, to minimize the difference between predicted and target receiver volumes. With viscosity and contact coefficients held fixed, we use MPM simulations to select the optimal angle for six target volumes ranging from $60$ to $160$~mL. The tilt duration follows a fixed commanded angular speed of $10^\circ$/s, while the hold and return commands are both $2$~s. We freeze the selected commands and predicted receiver volumes before performing five hardware repeats per target on the physical robot.

\textbf{Evaluation.} We evaluate pouring accuracy by reading the final volumes at the receiving cup across five trials for each of the six targets. The largest absolute difference between the mean received volume and the target is $3.8$~mL (Fig.~\ref{fig:pouring}b).
Although cup graduation accuracy and camera parallax remain uncalibrated, the resulting volumetric error remains low. This demonstrates that efficiently identifying the fluid's physical parameters from a single pour enables highly precise and repeatable manipulation.

\section{Conclusion}
We presented FORM, a weak-form approach to robotic manipulation of deformable materials that identifies material properties from observed motion and contact loads, without repeated simulation or gradient-based optimization during identification.
We use the same discretization for identification and forward MPM simulation, allowing the recovered model to transfer directly from a single probe to prediction and planning under new actions and geometries. Across elastic, elastoplastic, and fluid materials, results demonstrate that our approach enables efficient and reliable material identification for material-sensitive robotic manipulation.

\bibliographystyle{IEEEtran}
\bibliography{references}

\begin{thebibliography}{10}
\providecommand{\url}[1]{#1}
\csname url@rmstyle\endcsname
\providecommand{\newblock}{\relax}
\providecommand{\bibinfo}[2]{#2}
\providecommand\BIBentrySTDinterwordspacing{\spaceskip=0pt\relax}
\providecommand\BIBentryALTinterwordstretchfactor{4}
\providecommand\BIBentryALTinterwordspacing{\spaceskip=\fontdimen2\font plus
\BIBentryALTinterwordstretchfactor\fontdimen3\font minus
  \fontdimen4\font\relax}
\providecommand\BIBforeignlanguage[2]{{%
\expandafter\ifx\csname l@#1\endcsname\relax
\typeout{** WARNING: IEEEtran.bst: No hyphenation pattern has been}%
\typeout{** loaded for the language `#1'. Using the pattern for}%
\typeout{** the default language instead.}%
\else
\language=\csname l@#1\endcsname
\fi
#2}}

\bibitem{hansen2024td}
N.~Hansen, H.~Su, and X.~Wang, ``Td-mpc2: Scalable, robust world models for
  continuous control,'' in \emph{International Conference on Learning
  Representations}, vol. 2024, 2024, pp. 47\,376--47\,405.

\bibitem{khorrambakht2025worldplanner}
R.~Khorrambakht, J.~Ortiz-Haro, J.~Amigo, O.~Mostafa, D.~Dugas, F.~Meier, and
  L.~Righetti, ``Worldplanner: Monte carlo tree search and mpc with
  action-conditioned visual world models,'' \emph{arXiv preprint
  arXiv:2511.03077}, 2025.

\bibitem{thorpe2026physically}
A.~J. Thorpe, S.~Tretiakov, C.-H. Hsiao, S.~A. Low, X.~Li, H.~Iqbal, N.~P.
  Bhatt, U.~Topcu, and K.~Kumar, ``Physically viable world models: A case for
  query-conditioned embodied ai,'' \emph{arXiv preprint arXiv:2605.30542},
  2026.

\bibitem{ma2023learning}
P.~Ma, P.~Y. Chen, B.~Deng, J.~B. Tenenbaum, T.~Du, C.~Gan, and W.~Matusik,
  ``Learning neural constitutive laws from motion observations for
  generalizable pde dynamics,'' in \emph{International Conference on Machine
  Learning}.\hskip 1em plus 0.5em minus 0.4em\relax PMLR, 2023, pp.
  23\,279--23\,300.

\bibitem{li2023pac}
X.~Li, Y.-L. Qiao, P.~Y. Chen, K.~M. Jatavallabhula, M.~Lin, C.~Jiang, and
  C.~Gan, ``Pac-nerf: Physics augmented continuum neural radiance fields for
  geometry-agnostic system identification,'' \emph{arXiv preprint
  arXiv:2303.05512}, 2023.

\bibitem{yang2025differentiable}
X.~Yang, Z.~Ji, and Y.-K. Lai, ``Differentiable physics-based system
  identification for robotic manipulation of elastoplastic materials,''
  \emph{The International Journal of Robotics Research}, vol.~44, no.~13, pp.
  2126--2155, 2025.

\bibitem{pierron2012virtual}
F.~Pierron and M.~Gr{\'e}diac, \emph{The virtual fields method: extracting
  constitutive mechanical parameters from full-field deformation
  measurements}.\hskip 1em plus 0.5em minus 0.4em\relax Springer Science \&
  Business Media, 2012.

\bibitem{flaschel2021unsupervised}
M.~Flaschel, S.~Kumar, and L.~De~Lorenzis, ``Unsupervised discovery of
  interpretable hyperelastic constitutive laws,'' \emph{Computer Methods in
  Applied Mechanics and Engineering}, vol. 381, p. 113852, 2021.

\bibitem{flaschel2022discovering}
------, ``Discovering plasticity models without stress data,'' \emph{npj
  Computational Materials}, vol.~8, no.~1, p.~91, 2022.

\bibitem{hu20193}
Z.~Hu, T.~Han, P.~Sun, J.~Pan, and D.~Manocha, ``3-d deformable object
  manipulation using deep neural networks,'' \emph{IEEE Robotics and Automation
  Letters}, vol.~4, no.~4, pp. 4255--4261, 2019.

\bibitem{arriola2020modeling}
V.~E. Arriola-Rios, P.~Guler, F.~Ficuciello, D.~Kragic, B.~Siciliano, and J.~L.
  Wyatt, ``Modeling of deformable objects for robotic manipulation: A tutorial
  and review,'' \emph{Frontiers in Robotics and AI}, vol.~7, p.~82, 2020.

\bibitem{zhu2022challenges}
J.~Zhu, A.~Cherubini, C.~Dune, D.~Navarro-Alarcon, F.~Alambeigi, D.~Berenson,
  F.~Ficuciello, K.~Harada, J.~Kober, X.~Li, \emph{et~al.}, ``Challenges and
  outlook in robotic manipulation of deformable objects,'' \emph{IEEE Robotics
  \& Automation Magazine}, vol.~29, no.~3, pp. 67--77, 2022.

\bibitem{ying2024obstacle}
C.~Ying and K.~Yamazaki, ``Obstacle avoidance shape control of deformable
  linear objects with online parameters adaptation based on differentiable
  simulation,'' \emph{ROBOMECH Journal}, vol.~11, no.~1, p.~15, 2024.

\bibitem{shi2024robocraft}
H.~Shi, H.~Xu, Z.~Huang, Y.~Li, and J.~Wu, ``Robocraft: Learning to see,
  simulate, and shape elasto-plastic objects in 3d with graph networks,''
  \emph{The International Journal of Robotics Research}, vol.~43, no.~4, pp.
  533--549, 2024.

\bibitem{shi2023robocook}
H.~Shi, H.~Xu, S.~Clarke, Y.~Li, and J.~Wu, ``Robocook: Long-horizon
  elasto-plastic object manipulation with diverse tools,'' \emph{arXiv preprint
  arXiv:2306.14447}, 2023.

\bibitem{lin2022diffskill}
X.~Lin, Z.~Huang, Y.~Li, J.~B. Tenenbaum, D.~Held, and C.~Gan, ``Diffskill:
  Skill abstraction from differentiable physics for deformable object
  manipulations with tools,'' \emph{arXiv preprint arXiv:2203.17275}, 2022.

\bibitem{huang2023diffvl}
Z.~Huang, F.~Chen, Y.~Pu, C.~Lin, H.~Su, and C.~Gan, ``Diffvl: Scaling up soft
  body manipulation using vision-language driven differentiable physics,''
  \emph{Advances in Neural Information Processing Systems}, vol.~36, pp.
  29\,875--29\,900, 2023.

\bibitem{schenck2018spnets}
C.~Schenck and D.~Fox, ``{SPNets}: Differentiable fluid dynamics for deep
  neural networks,'' in \emph{Conference on Robot Learning}.\hskip 1em plus
  0.5em minus 0.4em\relax PMLR, 2018, pp. 317--335.

\bibitem{lopez2017adaptable}
T.~Lopez-Guevara, N.~K. Taylor, M.~U. Gutmann, S.~Ramamoorthy, and K.~Subr,
  ``Adaptable pouring: Teaching robots not to spill using fast but approximate
  fluid simulation,'' in \emph{Conference on Robot Learning}.\hskip 1em plus
  0.5em minus 0.4em\relax PMLR, 2017, pp. 77--86.

\bibitem{lin2022planning}
X.~Lin, C.~Qi, Y.~Zhang, Z.~Huang, K.~Fragkiadaki, Y.~Li, C.~Gan, and D.~Held,
  ``Planning with spatial-temporal abstraction from point clouds for deformable
  object manipulation,'' \emph{arXiv preprint arXiv:2210.15751}, 2022.

\bibitem{you2025make}
Y.~You, B.~Shen, C.~Deng, H.~Geng, S.~Wei, H.~Wang, and L.~Guibas, ``Make a
  donut: Hierarchical emd-space planning for zero-shot deformable manipulation
  with tools,'' \emph{IEEE Robotics and Automation Letters}, vol.~10, no.~4,
  pp. 3270--3277, 2025.

\bibitem{pmlr-v305-yamada25b}
J.~Yamada, S.~Zhong, J.~Collins, and I.~Posner, ``D-cubed: Latent diffusion
  trajectory optimisation for dexterous deformable manipulation,'' in
  \emph{Conference on Robot Learning}.\hskip 1em plus 0.5em minus 0.4em\relax
  PMLR, 2025, pp. 5039--5055.

\bibitem{ma2022risp}
P.~Ma, T.~Du, J.~B. Tenenbaum, W.~Matusik, and C.~Gan, ``Risp:
  Rendering-invariant state predictor with differentiable simulation and
  rendering for cross-domain parameter estimation,'' \emph{arXiv preprint
  arXiv:2205.05678}, 2022.

\bibitem{chen2026empm}
Y.~Chen, Y.~Hu, L.~Sun, T.~Kusnur, L.~Herlant, and C.~Jiang, ``Empm: Embodied
  mpm for modeling and simulation of deformable objects,'' \emph{IEEE Robotics
  and Automation Letters}, 2026.

\bibitem{zhang2024adaptigraph}
K.~Zhang, B.~Li, K.~Hauser, and Y.~Li, ``Adaptigraph: Material-adaptive
  graph-based neural dynamics for robotic manipulation,'' \emph{arXiv preprint
  arXiv:2407.07889}, 2024.

\bibitem{manoharan2025parameter}
N.~R. Manoharan, H.~Iqbal, and K.~Kumar, ``Parameter-efficient conditioning for
  material generalization in graph-based simulators,'' \emph{arXiv preprint
  arXiv:2511.05456}, 2025.

\bibitem{mei2019improving}
Y.~Mei and S.~Avril, ``On improving the accuracy of nonhomogeneous shear
  modulus identification in incompressible elasticity using the virtual fields
  method,'' \emph{International Journal of Solids and Structures}, vol. 178,
  pp. 136--144, 2019.

\bibitem{flaschel2023automated}
M.~Flaschel, S.~Kumar, and L.~De~Lorenzis, ``Automated discovery of generalized
  standard material models with euclid,'' \emph{Computer Methods in Applied
  Mechanics and Engineering}, vol. 405, p. 115867, 2023.

\bibitem{ingebrand2025basis}
T.~Ingebrand, A.~J. Thorpe, S.~Goswami, K.~Kumar, and U.~Topcu,
  ``Basis-to-basis operator learning using function encoders,'' \emph{Computer
  Methods in Applied Mechanics and Engineering}, vol. 435, p. 117646, 2025.

\end{thebibliography}

\end{document}